\documentclass[letterpaper]{article} % DO NOT CHANGE THIS
\usepackage{aaai2026}  % DO NOT CHANGE THIS
\usepackage{times}  % DO NOT CHANGE THIS
\usepackage{helvet}  % DO NOT CHANGE THIS
\usepackage{courier}  % DO NOT CHANGE THIS
\usepackage[hyphens]{url}  % DO NOT CHANGE THIS
\usepackage{graphicx} % DO NOT CHANGE THIS
\usepackage{natbib}  % DO NOT CHANGE THIS AND DO NOT ADD ANY OPTIONS TO IT
\usepackage{caption} % DO NOT CHANGE THIS AND DO NOT ADD ANY OPTIONS TO IT
\usepackage{algorithm}
\usepackage{algorithmic}

\usepackage{booktabs}
\usepackage{multirow}
\usepackage{graphicx}
\usepackage{amsfonts}
\usepackage{amsmath}
\usepackage{xcolor}

\usepackage{amssymb}
\usepackage{pifont}
\newcommand{\xmark}{\ding{55}}
\usepackage{newfloat}
\usepackage{listings}
\DeclareCaptionStyle{ruled}{labelfont=normalfont,labelsep=colon,strut=off} % DO NOT CHANGE THIS
\floatstyle{ruled}
\newfloat{listing}{tb}{lst}{}
\floatname{listing}{Listing}
\title{Where It Moves, It Matters: Referring Surgical Instrument Segmentation via Motion}
\author{
    Meng Wei\textsuperscript{\rm 1},
    Kun Yuan\textsuperscript{\rm 2,3,5}\footnotemark[1],
    Shi Li\textsuperscript{\rm 3},
    Yue Zhou\textsuperscript{\rm 2},
    Long Bai\textsuperscript{\rm 4},
    Nassir Navab\textsuperscript{\rm 2},
    Hongliang Ren\textsuperscript{\rm 4},\\
    Hong Joo Lee\textsuperscript{\rm 2,5}\thanks{Corresponding Author.},
    Tom Vercauteren\textsuperscript{\rm 1}\thanks{These authors contributed equally.},
    Nicolas Padoy\textsuperscript{\rm 3}\footnotemark[2]
}

\affiliations{
    \textsuperscript{\rm 1}Biomedical Engineering and Imaging Sciences, King's College London\\
    \textsuperscript{\rm 2}Technical University of Munich\\
    \textsuperscript{\rm 3}ICube, University of Strasbourg\\
    \textsuperscript{\rm 4}Department of Electrical Engineering, The Chinese University of Hong Kong \\
    \textsuperscript{\rm 5} Munich Center for Machine Learning \\
    \{meng.wei, tom.vercauteren\}@kcl.ac.uk,
    \{yue.zhou, nassir.navab, hongjoo.lee, kun.yuan\}@tum.de \\
    shi.li@ext.ihu-strasbourg.eu, \{b.long@link, hlren@ee\}.cuhk.edu.hk, 
    npadoy@unistra.fr
}
\usepackage{bibentry}
\begin{document}

\maketitle

\begin{abstract}
Enabling intuitive, language-driven interaction with surgical scenes is a critical step toward intelligent operating rooms and autonomous surgical robotic assistance. However, the task of referring segmentation, localizing surgical instruments based on natural language descriptions, remains underexplored in surgical videos, with existing approaches struggling to generalize due to reliance on static visual cues and predefined instrument names. In this work, we introduce \textit{SurgRef}, a novel motion-guided framework that grounds free-form language expressions in instrument motion, capturing how tools move and interact across time, rather than what they look like. This allows models to understand and segment instruments even under occlusion, ambiguity, or unfamiliar terminology. To train and evaluate SurgRef, we present \textit{Ref-IMotion}, a diverse, multi-institutional video dataset with dense spatiotemporal masks and rich motion-centric expressions. SurgRef achieves state-of-the-art accuracy and generalization across surgical procedures, setting a new benchmark for robust, language-driven surgical video segmentation.

\end{abstract}


\section{Introduction}

Surgical data science presents a critical and high-stakes application field for computer vision, where the model's performance has a direct impact on patient safety and clinical outcomes. A core challenge in this field is the analysis of endoscopic video data, which requires recognizing surgical instruments~\cite{twinanda2016endonet,allan20192017,allan20202018}, anatomical structures~\cite{carstens2023dresden}, and procedural actions~\cite{nwoye2021rendezvous}. Such capabilities enable real-time comprehension of surgical workflows, support skill assessment~\cite{gupta2024multi}, and facilitate complication prediction~\cite{ma2022surgical} and robotic navigation~\cite{long2025surgical}. Central to these functions is the task of surgical video object segmentation, which provides precise, pixel-level identification of surgical instruments and anatomies. Despite the progress, recent studies have demonstrated that models pretrained on general-purpose datasets~\cite{ding2023mevis,cheng2022masked,huang2025segment} often fail to generalize in surgical settings due to domain-specific complexities such as visual occlusion, similar-looking tools, and challenging imaging conditions, e.g., bleeding, smoke, and poor lighting~\cite{nwoye2025cholectrack20,sun2023cifar,sun2024assessing}. This highlights the importance of surgical domain-specific datasets and methods tailored to the unique visual and temporal characteristics of surgical scenes. 

In this work, we move beyond conventional surgical segmentation \cite{wei2025segmatch} by tackling surgical referring segmentation, a task tailored to the surgical domain where instruments are localized based on static language cues~\cite{wang2024video,liu2025resurgsam2} such as appearance and spatial context (e.g., ``the grasper on the left''), and dynamic motion-centric language expressions (e.g., ``the tool grasping gallbladder''), as illustrated in Figure~\ref{fig:data}. This enables intuitive interaction between surgeons and AI systems. When integrated with AR, it supports surgical training by allowing users to query instruments or actions using language commands~\cite{killeen2024take}. Intraoperatively, it acts as an intelligent assistant, highlighting key targets, answering verbal queries, and guiding surgeons~\cite{seenivasan2022surgical,yuan2024advancing,huang2025surgtpgs}. Also, it supports natural human-robot collaboration through language-based procedural commands in robotic surgery (e.g., ``cut where the grasper is holding''), allowing for fine-grained semantic control for surgery automation~\cite{long2025surgical}.

However, enabling robust referring segmentation in surgical settings presents unique visual and linguistic challenges. Specifically, surgical scenes are complex and dynamic due to the visual occlusions, deformable anatomical structures, and the simultaneous presence of multiple similar-looking instruments. These factors introduce significant ambiguity when localizing target objects based solely on visual appearance. Linguistically, surgical language expressions are highly specialized and lack standardization across institutions and regions~\cite{lavanchy2024challenges}. The same instrument may be described in different ways depending on local surgical protocols, which limits the generalization capability of existing language-conditioned models~\cite{yuan2025recognizing,yuan2025learning}. Existing methods often depend on static language expressions~\cite{wang2024video}, such as explicit instrument names (e.g., ``Prograsp forceps'') for grounding, which restricts their robustness in handling unseen procedures or adapting to different clinical centers.

In contrast, instrument motion provides a more consistent and interpretable signal. Surgical procedures are defined by sequences of tool actions, including entry paths, retraction patterns, and tool-tissue interactions. These motions are important to both procedural understanding and surgical training. For example, the process of achieving the Critical View of Safety (CVS) in laparoscopic cholecystectomy~\cite{murali2022latent} involves a well-defined series of instrument gestures that experienced surgeons can consistently recognize, regardless of where they were trained.

Therefore, we propose SurgRef, a motion-guided referring video segmentation framework that grounds natural language expressions in surgical video by explicitly modeling instrument motion. In addition to existing methods that rely on static appearance cues or explicit instrument names, SurgRef also interprets motion-centric expressions (e.g., ``the tool entering from the right and retracting the gallbladder medially'') to produce fine-grained, temporally localized segmentations. To support this, we construct the Ref-IMotion dataset, which aggregates three public datasets to cover diverse surgical procedures and settings: laparoscopic cholecystectomy from Cholec80~\cite{twinanda2016endonet}, robot-assisted porcine surgery from EndoVis~\cite{allan20192017}, and robotic-assisted laparoscopic radical prostatectomy from GraSP~\cite{ayobi2024pixel}. Its composition spans both laparoscopic and robotic modalities, sourced from different institutions, enabling a comprehensive benchmark for evaluating model generalization across varied clinical domains and toolsets. For datasets lacking natural language supervision, we manually annotate high-quality motion-centric referring language expressions that describe instrument behavior such as entry trajectories, retraction patterns, and tool-tissue interactions (e.g., ``the tool that enters from the top and pulls the gallbladder to the left''). The dataset includes 319 surgical video clips with a total of 21,350 annotated frames and 718 referring expressions, including 358 motion-based expressions, each paired with dense frame-level masks and temporally localized intervals.

In this work, SurgRef enhances temporal reasoning by our proposed key-frame attention module that adaptively selects a subset of expression-relevant frames by leveraging language-guided object semantics. Instead of uniformly processing all frames, the module computes frame-wise relevance scores from decoder-level object queries conditioned on the referring expression. By selecting the top-ranked frames for segmentation, this module enables the model to focus on temporally salient moments, suppressing redundancy and emphasizing frames with meaningful object interactions. This motion-centric design improves both temporal efficiency and the precision of spatial-temporal grounding in long surgical videos. SurgRef segments the spatial regions of referred objects while identifying expression-relevant frames through language-guided key-frame selection, achieving state-of-the-art performance in surgical video referring segmentation. Its motion-guided representations also exhibit strong generalization across diverse surgical procedures and toolsets. {Our main contributions are summarized as follows:
\begin{itemize}
    \item We construct the Ref-IMotion dataset by annotating multiple public datasets with diverse surgical procedures, and manually curating motion-based referring expressions paired with dense spatial-temporal masks.

    \item We propose SurgRef, a motion-guided referring video segmentation framework that grounds both static and motion-centric language expressions in surgical videos, enabling fine-grained spatial-temporal understanding.    
    
    \item Extensive experiments demonstrate that SurgRef achieves state-of-the-art performance and strong generalization across varied surgical toolsets and procedures, setting a robust benchmark for motion-guided video segmentation in surgery.
\end{itemize}

\section{Related Work}
\paragraph{Video Segmentation.}
Video segmentation aims to assign pixel-wise labels to each frame in a video sequence, enabling the consistent tracking and delineation of objects or regions of interest over time \cite{oh2019video,yang2021associating,cheng2022xmem,yang2022decoupling}. Unlike image segmentation, it accounts for temporal continuity, motion dynamics, occlusions, and appearance variations, making it a more challenging task. Since it plays a critical role in various applications, such as autonomous driving \cite{siam2018comparative,kim2020highway,muhammad2022vision,sun2023alice} and surgical scene understanding \cite{ni2020attention,ayobi2023matis,visnet,zhou2023text}, extensive studies have been conducted by exploiting spatiotemporal feature modeling, memory-based architectures, and attention mechanisms to improve temporal consistency and accuracy across frames. For example, Oh et al. \cite{oh2019video} proposed STM. They introduced a memory network that retrieves useful information from past frames to guide segmentation in the current frame, improving temporal consistency. Also, Cheng et al. \cite{cheng2022xmem} proposed XMem. It extends the STM by using a dynamic memory update strategy that allows scalable and efficient segmentation over long video sequences. Both methods highlight the effectiveness of memory-based architectures for robust video object segmentation.

\begin{figure*}[htbp]
    \centering
    \includegraphics[width = \linewidth]{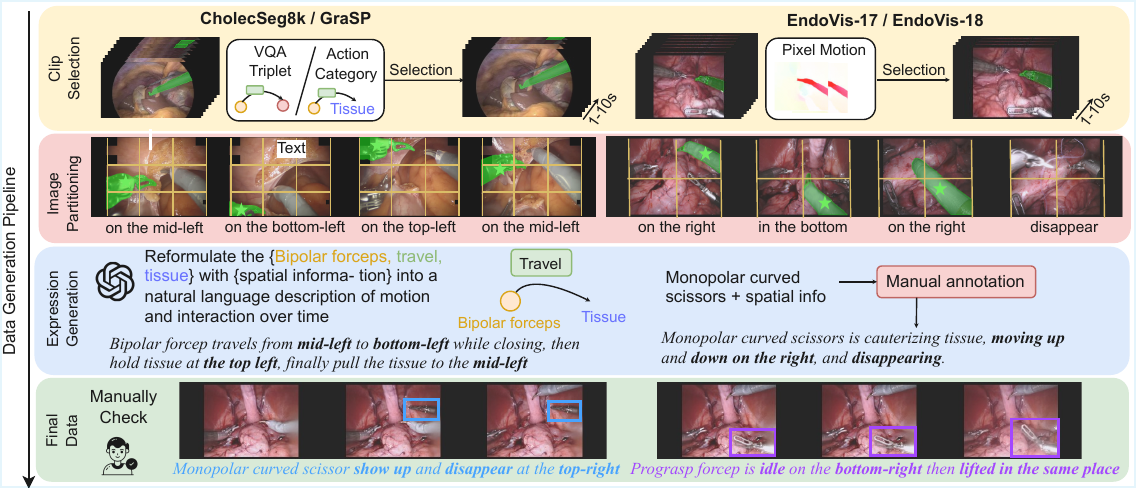}
    \caption{Data construction process of the EndoVis-IM17, EndoVis-IM18, GraSP-IM, and CholecSeg8k-IM.}
    \label{fig:data}
\end{figure*}

\subsubsection{Surgical Video Segmentation.}
Surgical video segmentation has been extensively studied as a core task in computer-assisted and robotic surgery. It serves as a foundation for a variety of downstream applications, including tool tracking, workflow analysis, surgical skill assessment, and autonomous navigation. Most prior work has focused on vision-based models trained with pixel-level supervision \cite{ni2020attention,ayobi2023matis,ayobi2024pixel,grammatikopoulou2024spatio,liao2025disentangling}. For example, Ni et al.\cite{ni2020attention} introduced LWANet, an attention-guided lightweight architecture that employs depth-wise separable convolutions to enable real-time inference with reduced computational cost. Ayobi et al.~\cite{ayobi2023matis} proposed MATIS, a transformer-based framework that combines pixel-wise attention, masked attention modules focused on instrument regions, and video transformers for handling short-range temporal dependencies.

\subsubsection{Surgical Video Referring Segmentation.}
Beyond conventional vision-based models, vision-language approaches is the new trend for surgical instrument segmentation because of their human-understandable interactions. Wang et al.\cite{visnet} introduced VIS-Net, the first method for referring instrument segmentation in the surgical domain. Their approach incorporates a video-instrument synergistic network to jointly learn video-level and instrument-level representations. Zhou et al.\cite{zhou2023text} proposed TP-SIS, a text-promptable segmentation framework that integrates pretrained vision-language models and utilizes a mixture-of-prompts mechanism to improve robustness.

Despite recent progress, existing methods fall short in two key aspects: they lack long-term temporal modeling and overly rely on static appearance cues. This overlooks motion, the most consistent and discriminative signal in surgical scenes, especially when tools are visually similar or occluded. Without motion-aware modeling, current approaches struggle to robustly localize and segment instruments in complex, real-world surgical scenarios.

\paragraph{Surgical Instrument Segmentation Dataset.}
Several benchmarks have been developed for surgical instrument segmentation, including EndoVis-17, EndoVis-18, Cholec80~\cite{hong2020cholecseg8k} and GraSP \cite{ayobi2024pixel}. The EndoVis-17 dataset consists of stereo video sequences acquired from the da Vinci robotic system, annotated with binary, part-based, and type-level segmentation masks for multiple surgical instruments. EndoVis-18 extends this to full-scene segmentation, providing labels for both instruments and anatomical structures across longer and more varied procedures. CholecSeg8k is a semantic segmentation dataset derived from the Cholec80 surgical video benchmark~\cite{twinanda2016endonet}, offering dense annotations for surgical instruments. GraSP is a benchmark dataset designed to support surgical scene understanding through a hierarchy of complementary tasks at varying levels of granularity. It includes pixel-level instance annotations, as well as instrument and action category labels.

These datasets lack fine-grained motion annotations, only providing object identity, spatial location, and short-term action descriptions without describing the trajectory, direction, and motion of instruments. This limits their applicability for training or evaluating models that aim to understand and leverage motion cues in referring segmentation tasks.

\begin{figure*}[!htbp]
    \centering
    \includegraphics[width=\linewidth]{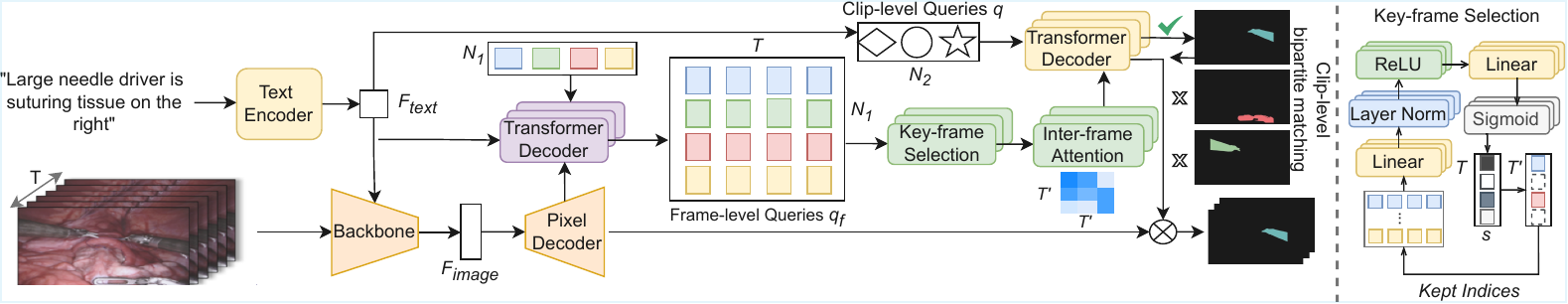}
    \caption{Baseline model for SurgRef, which integrates a Swin Transformer backbone for spatial feature extraction, a frozen RoBERTa encoder for language grounding, a language-guided transformer decoder for cross-modal reasoning, and a key-frame attention module for selecting semantically aligned frames.}
    \label{fig:model}
\end{figure*}

\section{Methodology}
\subsection{Dataset Construction}
We tackle the task of segmenting surgical objects in a video guided by motion-centric language expressions. To this end, we construct the motion-guided referring dataset, i.e., Ref-IMotion, by annotating publicly available surgical videos, EndoVis-17, EndoVis-18, CholecSeg8k, and GraSP, with rich motion-based textual descriptions. 

For each dataset, we follow its own preprocessing pipelines to extract high-quality, mask-annotated frames. From these, we select clips that exhibit high motion complexity, including instrument appearances, disappearances, and interactions. Specifically, we extract frames from EndoVis-17, EndoVis-18, and GraSP where tool interactions are detected via abrupt pixel intensity changes or rapid motion. For CholecSeg8k, we use scene graph triplets from SSG-VQA \cite{yuan2024advancing} to locate frames with distinct tool-object interactions. Frames are grouped into short clips based on visual continuity and action diversity, resulting in a curated set of 319 video clips totaling 21{,}350 frames, the largest video referring segmentation dataset in surgical computer vision to date.

Our core contribution lies in introducing motion-centric expressions that capture not only the presence and action of surgical instruments, but also their spatiotemporal dynamics. To reflect this focus on motion, we prefix our newly annotated datasets with ``IM."
\begin{itemize}
    \item For EndoVis-IM17 (EV-IM17), we annotate expressions aligned with motion trajectories across frame sequences (e.g., ``Grasping retractor appears from the left, moves to the top-right, and disappears''), supplemented with coarse spatial references using a $2\times2$ grid.
    \item EndoVis-IM18 (EV-IM18) features attribute-based motion expressions (e.g., ``Suction instrument is suctioning blood'') that focus on tool action and spatial movement.
    \item For CholecSeg8k-IM (C8K-IM), we augment existing segmentation masks with semantic triplets and $3\times3$ grid-based spatial tags to construct expressions such as ``Grasper is grasping gallbladder in the mid-left.'' While some triplets are adapted from SSG-VQA \cite{yuan2024advancing}, we perform manual verification and correction to ensure consistency and domain accuracy.
    \item To construct GraSP-IM (GSP-IM), we first selected surgical videos based on specific procedure phases and steps, assigning each video a unique identifier in the format \textit{case-phase-step}. Using the previously defined action categories, we generated a corresponding motion expression for each object in each video segment. Object instances were identified from the annotations, and their bounding boxes were extracted accordingly. The key point of each object was defined as the center of its bounding box. To estimate motion direction, we calculated the displacement of this point across consecutive frames. Each frame was divided into a $3\times3$ spatial grid, allowing us to localize movements within spatial regions and combine them with action labels. We then used a prompt to feed the structured inputs, comprising instrument, action, object, and spatial information, into a large language model \cite{openai2025chatgpt} to generate natural language descriptions of motion and interaction over time. Note that we merged video segments into their neighboring steps if the number of images was fewer than 10, resulting in 258 unique videos and 10{,}944 frames in total.
\end{itemize}
% For EndoVis-IM17, we annotate expressions aligned with motion trajectories across frame sequences (e.g., ``Grasping retractor appears from the left, moves to the top-right, and disappears''), supplemented with coarse spatial references using a $2\times2$ grid. EndoVis-IM18 features attribute-based expressions (e.g., ``Suction instrument is suctioning blood'') that focus on tool function without explicit spatial or temporal cues. For CholecSeg8k, we augment existing segmentation masks with semantic triplets and $3\times3$ grid-based spatial tags to construct expressions such as ``Grasper is grasping gallbladder in the mid-left.'' While some triplets are adapted from SSG-VQA, we perform manual verification and correction to ensure consistency and domain accuracy. For GraSP
Together, these curated annotations form the existing publicly available datasets that compose our Ref-IMotion, supporting fine-grained spatio-temporal language grounding and advancing motion-aware surgical scene understanding.
\begin{table*}[htbp]
\centering
\setlength{\tabcolsep}{4pt}
\begin{minipage}{0.39\textwidth}
\centering
\resizebox{\textwidth}{!}{
\begin{tabular}{lcccc}
\hline
Dataset & Static Attr. & Motion& Total & Motion \% \\
\hline
EndoVis-IM17 &  119 & 45 & 164 & 27.43 \\
EndoVis-IM18 & 137 & 21 & 158 & 13.29 \\
CholecSeg8k-IM & 104 & 34 & 138 & 24.63 \\
GraSP-IM      & N/A & 258 & 258 & 100 \\
\midrule
Ref-IMotion (total)   & 360   & 358 & 718 & 49.86 \\
\hline
\end{tabular}
}
\end{minipage}
\hfill
\begin{minipage}{0.6\textwidth}
\centering
\resizebox{\textwidth}{!}{
\begin{tabular}{lccccc}
\hline
Dataset & Video-based & TC & RE & Inst.M & SE \\
\hline
EndoVis-RS17/ 18 \cite{wang2024video} & \checkmark & \xmark & \checkmark & \checkmark & RGB color$^a$ \\
CholecSeg8k \cite{hong2020cholecseg8k}  & \xmark & \xmark & \xmark & \checkmark & \xmark \\
SSG-VQA \cite{yuan2024advancing} & \xmark & \xmark & Triplet & \xmark & \xmark \\
\hline
Ref-IMotion   & \checkmark & \checkmark & \checkmark & \checkmark & 2$\times$2 / 3$\times$3$^b$ \\
\hline
\end{tabular}
}
\end{minipage}
\caption{Left: Ref-IMotion expression statistics (Attr.: Attribute). Right: Comparison of multimodal-temporal properties with other surgical datasets (TC: Temporal Continuity; RE: Referring Expression; Inst.M: Instance Mask; SE: Spatial encoding). $^a$Color-coded segmentation masks for instrument instances. $^b$Text-based spatial grid descriptions for instrument positioning.}
\label{tab:dataset}
\end{table*}

Each surgical video consists of $T$ RGB frames, denoted as $\{I_t\}_{t=1}^{T}$, where each frame $I_t \in \mathbb{R}^{H \times W \times 3}$ with spatial resolution $H \times W$. To extract spatial features from each frame, we employ a vision transformer backbone based on the Swin Transformer~\cite{liu2021swin}, yielding a multi-scale feature representation for each frame. Let $F_{image}$ denote the extracted visual features from frame $I_t$. 

Given a natural language expression that refers to surgical activity (e.g., \textit{``Grasper is retracting tissue on the right''}), our goal is to perform referring video object segmentation: producing a segmentation mask for the object(s) referred to by the expression in each frame. To interpret the referring expression, we use a frozen RoBERTa-base~\cite{liu2019roberta} model as our text encoder. This encoder transforms the unstructured language input into a dense semantic embedding. Let $F_{\text{text}} \in \mathbb{R}^{d}$ denote the global sentence-level embedding extracted from RoBERTa.

Unlike the object queries in mask2former \cite{cheng2022masked}, SurgRef initializes $N_2$ language-driven queries $\{q_i^{(0)}\}_{i=1}^{N_2}$ by injecting the language embedding $F_{\text{text}}$ into the initial query vectors:

\begin{equation}
q_i^{(0)} = W_{\text{init}} \cdot F_{\text{text}} + b_i,
\end{equation}

where $W_{\text{init}} \in \mathbb{R}^{d_q \times d}$ is a learnable projection matrix and $b_i \in \mathbb{R}^{d_q}$ is a learnable positional bias for the $i$-th query. Our transformer decoder is based on the Mask2Former~\cite{cheng2022masked} architecture, adapted to incorporate language guidance. Specifically, we first replace standard object queries with language-driven initializations, then inject textual semantics into the cross-attention layers for spatial reasoning. After $L$ transformer decoder layers, the final query representations $\{q_i^{(L)}\}_{i=1}^{N_2}$ are passed through two parallel heads:

\paragraph{Classification Head.} Each query produces a classification score indicating whether it corresponds to a relevant object:

\begin{equation}
\hat{y}_i^{\text{cls}} = \text{Linear}\left( \left[ q_i^{(L)} \, \| \, F_{\text{text}} \right] \right) \in \mathbb{R}^{C + 1},
\end{equation}

where $\|$ denotes vector concatenation, and $C$ is the number of object classes (with one additional background class).

\paragraph{Mask Embedding Head.} Each query also generates a mask embedding:

\begin{equation}
\hat{y}_i^{\text{mask}} = \text{MLP}(q_i^{(L)}) \in \mathbb{R}^{d_m}.
\end{equation}

Given the pixel-level mask features $F_{\text{mask}} \in \mathbb{R}^{d_m \times H \times W}$, the binary segmentation mask $\hat{M}_i$ for query $i$ is computed by a dot product and sigmoid activation:

\begin{equation}
\hat{M}_i = \sigma\left( e_i^\top \cdot F_{\text{mask}} \right),
\end{equation}

where $e_i = \hat{y}_i^{\text{mask}}$ is the $i$-th mask embedding, and $\sigma$ denotes the sigmoid function. To determine valid object-mask pairs, we apply a threshold on classification confidence. All queries with $\max_{c=1}^{C} \text{softmax}(\hat{y}_i^{\text{cls}})_c > \tau$ (where $\tau = 0.8$) are retained. This thresholding strategy allows the model to flexibly handle both single-object and multi-object expressions, which is essential for parsing complex surgical scenes.

\paragraph{Key-frame Selection.}
Surgical videos contain significant temporal redundancy, with many frames showing repetitive activity. To improve temporal reasoning and reduce computational cost, we introduce a key-frame selection module that selects expression-relevant frames using language-guided queries.
% For each frame $I_t$, we extract its language-guided object representation $\mathbf{e}_t$ by aggregating the corresponding object queries after the transformer decoder. We then compute a scalar relevance score $s_t \in [0, 1]$ using a lightweight attention MLP applied to $\mathbf{e}_t$:

% \begin{equation}
% s_t = \sigma\left( \mathbf{W}_2 \cdot \text{ReLU} \left( \mathbf{W}_1 \cdot \mathbf{e}_t \right) + \mathbf{b} \right),
% \end{equation}

% where $\mathbf{W}_1$, $\mathbf{W}_2$ are learnable projection matrices, $\mathbf{b}$ is a bias vector, and $\sigma(\cdot)$ denotes the sigmoid function. The embedding $\mathbf{e}_t$ can be obtained by averaging the final object queries corresponding to frame $t$, optionally weighted by classification confidence.

% After computing scores $\{s_t\}_{t=1}^{T}$ for all frames, we select the top-$T'$ frames (e.g., $T'=32$ or $T'=16$) with the highest relevance scores. Only these key frames are passed into the segmentation decoder for mask prediction. This strategy allows the model to focus computation on frames that are semantically aligned with the referring expression, improving both efficiency and expression-specific segmentation performance in long surgical sequences.

For each video clip, the transformer decoder generates language-guided frame-level queries $\mathbf{q}_f \in \mathbb{R}^{C_Q \times N_1}$ of dimension $C_Q$ from the transformer decoder, which incorporate both spatial and linguistic context, yielding a tensor of size $[T, N_1, C_Q]$.
For each frame $I_t$, we extract its frame-level representation $\mathbf{e}_t \in \mathbb{R}^{C_Q}$ by aggregating the corresponding object queries, resulting in embeddings of size $[T, C_Q]$.
We then compute a scalar relevance score $s_t \in [0, 1]$ for each frame using a lightweight MLP applied to $\mathbf{e}_t$:
\begin{equation}
s_t = \sigma\left( \mathbf{W}_2 \cdot \text{ReLU} \left( \mathbf{W}_1 \cdot \mathbf{e}_t \right) + \mathbf{b} \right),
\end{equation}
where $\mathbf{W}_1 \in \mathbb{R}^{d \times C_Q}$, $\mathbf{W}_2 \in \mathbb{R}^{1 \times d}$ are learnable projection matrices, $\mathbf{b}$ is a bias term, and $\sigma(\cdot)$ denotes the sigmoid function.
These scores measure the visual-text alignment between each frame and the referring expression.

After computing relevance scores $\{s_t\}_{t=1}^{T}$ for all frames, we select the top-$T'$ frames with the highest scores, maintaining temporal order.
Only these key frames are passed to the segmentation decoder for final mask prediction.
For example, for the expression ``scissors traveling,'' frames showing active motion score $>0.8$, while idle frames score $<0.3$. This strategy allows the model to focus computation on frames that are semantically aligned with the referring expression, improving both efficiency and expression-specific segmentation performance in long surgical sequences.
\paragraph{Inter-frame Attention.}
After key-frame selection, we apply the inter-frame attention~\cite{ding2023mevis} to aggregate temporal motion information across the selected key frames. The language-guided frame-level queries $\mathbf{q}_f$ now yield a tensor of size $[T', N_1, C_Q]$, where $T'$ denotes the number of selected key frames and $N_1$ is the number of queries per frame. We flattend $\mathbf{q}_f$ into $\mathbf{q}_f' \in \mathbb{R}^{(T' \cdot N_1) \times C_Q}$ then apply multi-head self-attention followed by feed-forward networks with residual connections, which enables queries to exchange information across the temporal dimension, capturing object motion. Then $\mathbf{q}_f'$ is reshaped back and passed to the transformer decoder for mask generation.
\paragraph{Training Loss.} The model is trained with a composite loss that combines frame-level loss, which includes classification loss (cross-entropy) and mask prediction losses (binary cross-entropy and Dice), and video-level losses, including temporal similarity loss that enforces consistency of query embeddings across adjacent frames and video-level mask prediction loss (see supplementary).

\begin{table}[htbp]
\centering
\resizebox{\linewidth}{!}{
\begin{tabular}{llccccc}
\toprule
Dataset & Model & J & F & J\&F & Dice & IoU \\
\midrule
\multirow{4}{*}{EV-IM17} 
  % & VIS-Net \cite{wang2024video}
 & VIS-Net & 79.21 & 82.52 & 80.86 & 82.90 & 78.24 \\
  % & VISA \cite{yan2024visa} 
  & VISA& 86.85 & 84.93 & 85.89 & 83.39 & 81.57 \\
  % & MPG-SAM 2 \cite{rong2025mpg} 
  & MPG-SAM 2 & 88.20 & \textbf{90.61} & 89.41 & 84.37 & \textbf{82.94} \\
  \cmidrule{2-7}
  & Ours & 87.11 & 88.29 & 87.70 & 83.14 & 80.16 \\
  & Ours + KFS & \textbf{89.91} & 88.93 & \textbf{89.42} & \textbf{87.94} & 82.12 \\
\midrule
\multirow{4}{*}{EV-IM18} 
  % & VIS-Net \cite{wang2024video}
  & VIS-Net & 58.51 & 59.56 & 59.03 & 54.24 & 49.97 \\
  % & VISA \cite{yan2024visa} 
  & VISA & 83.46 & 81.96 & 82.71 & 63.77 & 58.72 \\
  % & MPG-SAM 2 \cite{rong2025mpg} 
  & MPG-SAM 2 & 83.11 & 82.94 & 83.03 & 62.14 & 59.67 \\
  \cmidrule{2-7}
  & Ours & 80.08 & 80.45 & 80.27 & 57.86 & 55.23 \\
  & Ours + KFS & \textbf{85.91} & \textbf{83.05} & \textbf{84.48} & \textbf{66.89} & \textbf{61.72} \\
\midrule
\multirow{2}{*}{GSP-IM}
  & Ours & 81.36 & 81.67 & 81.52 & 70.63 & 66.92 \\
  & Ours + KFS & \textbf{84.92} & \textbf{85.01} & \textbf{84.97} & \textbf{73.56} & \textbf{70.07} \\
\midrule
\multirow{2}{*}{C8K-IM*} 
  & Ours & 64.58 & 71.23 & 67.91 & 67.89 & 63.54 \\
  & Ours + KFS & \textbf{66.92} & \textbf{74.31} & \textbf{70.62} & \textbf{68.13} & \textbf{66.23} \\
\bottomrule
\end{tabular}}
\caption{Comparison of state-of-the-art methods, our baseline model adapted from LMPM~\cite{ding2023mevis}, and our full approach with the key-frame selection (KFS) module. Models are trained and tested separately on EndoVis-IM17, EndoVis-IM18, and GraSP-IM datasets. CholecSeg8k-IM is an unseen test set, where SurgRef is trained on GraSP-IM and evaluated zero-shot on CholecSeg8k-IM.}
\label{tab1}
\end{table}

% \begin{table*}[htbp]
% \centering
% \small
% \begin{tabular}{llcccccc}
% \toprule
% \multirow{2}{*}{\textbf{Dataset}} & \multirow{2}{*}{\textbf{Testing Expression}} 
% & \multicolumn{3}{c}{\textbf{w/o Motion}} 
% & \multicolumn{3}{c}{\textbf{w/ Motion}} \\
% \cmidrule(lr){3-5} \cmidrule(lr){6-8}
% & & \textbf{J} & \textbf{F} & \textbf{J\&F} & \textbf{J} & \textbf{F} & \textbf{J\&F} \\
% \midrule

% \multirow{3}{*}{EndoVis-IM17} 
% & Static features            & 73.92 & 74.58  & 74.25 & 85.42 & 84.91 & 85.17 \\
% & w/ Local information       & 76.33 & 76.95    & 76.64 & 87.52 & 87.13 & 87.33 \\
% & w/ Motion-based expression & 79.92 & 78.14    & 79.03   & 89.91 & 88.93 & 89.42 \\
% \midrule

% \multirow{3}{*}{EndoVis-IM18} 
% & Static features            & 67.78 & 68.53 & 68.16 & 82.77 & 79.26 & 81.02 \\
% & w/ Local information       & 69.24 & 69.05 & 69.15 & 84.13 & 81.97 & 83.05 \\
% & w/ Motion-based expression & 71.79 & 70.32 & 71.06 & 85.91 & 83.05 & 84.48 \\
% \midrule

% \multirow{3}{*}{GraSP-IM} 
% & Static features            & 68.01 & 68.75 & 68.38 & 81.02 & 81.93 & 81.48 \\
% & w/ Local information       & 68.92 & 69.88 & 69.40 & 81.53 & 82.74 & 82.14 \\
% & w/ Motion-based expression & 70.18 & 71.34 & 70.76 & 84.92 & 85.01 & 84.97 \\
% \bottomrule

% \end{tabular}
% \caption{Comparison of model performance across datasets evaluating with different expression styles. Each test input is evaluated using our model trained with (w/) or without (w/o) motion-aware expressions.}
% \label{tab3}
% \end{table*}
\section{Experiments}
\subsection{Evaluation Metrics}
We adopt J, F, J\&F, Dice, and IoU as evaluation metrics. Our surgical referring video segmentation model requires tracking referred objects across the entire video, even under occlusion. Therefore, to capture both spatial and boundary accuracy, we use J (region similarity) and F (contour accuracy), with J\&F denoting their mean. All metrics are computed frame-wise between predicted and ground-truth masks, averaged temporally per object, and then aggregated across all referred objects. We also report Dice and IoU for pixel-wise overlap quality. Details are in the supplementary. 

\subsection{Implementation Details}
We use a Swin Transformer~\cite{liu2021swin} as the visual backbone and a frozen RoBERTa-base~\cite{liu2019roberta} model as the textual encoder. We set $T'$ to 8 for the key-frame selection module for optimal efficiency-accuracy trade-off. The model is trained for 100{,}000 iterations using the AdamW optimizer with an initial learning rate of $5 \times 10^{-5}$, decayed by a cosine scheduler. The batch size is set to 8 and the weight decay to 0.05. Training takes approximately 23 hours on two NVIDIA A100 GPUs with 40 GB of memory. Implementation details are in the supplementary.

\subsection{Dataset Details}

Table~\ref{tab:dataset} presents a comprehensive comparison of Ref-IMotion with existing referring video segmentation datasets in the general computer vision and surgical computer vision fields. It analyzes these datasets across multiple dimensions, including input modality, temporal reasoning, referring expression types, and spatial annotations. Overall, Ref-IMotion is the only dataset that integrates all these features, which enable fine-grained spatiotemporal grounding. Prior referring datasets like Ref-COCO and Ref-Youtube-VIS support temporal language grounding but are built on natural scenes and lack the visual complexity of surgical domains. Surgical datasets such as EndoVis-IM17/18 and CholecSeg8k-IM provide valuable surgical segmentation masks but contain either limited or no referring language expression, restricting their utility for motion-aware language-driven tasks. Our proposed Ref-IMotion dataset with carefully annotated motion-centric expressions that describe not just what an instrument is, but how it moves through space and time, captures spatiotemporal dynamics such as entry trajectories, retraction paths, and tool-tissue interactions. For instance, our EndoVis-IM17 subset includes trajectory-aligned motion expressions mapped to frame sequences, while the CholecSeg8k-IM subset includes expressions enriched with semantic action triplets and grid-based spatial tags. Unlike attribute-only phrases (e.g., ``suctioning blood''), our expressions support dynamic interaction grounding (e.g., ``grasper enters from the left and retracts the gallbladder medially''), which is crucial for real-world surgical understanding.

\begin{figure}[!b]
    \centering
    \includegraphics[width = \linewidth]{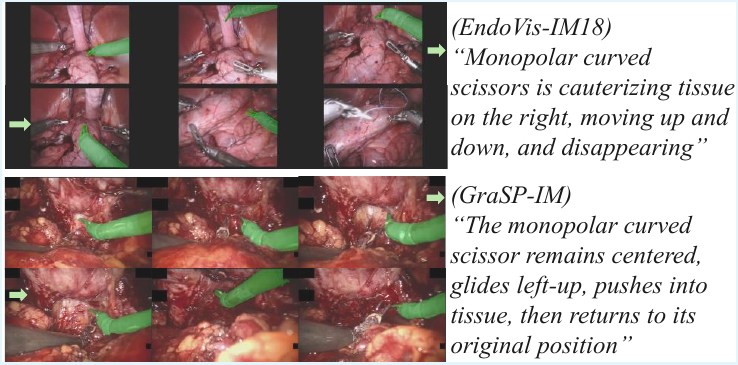}
    \caption{Visualization examples of our SurgRef on EndoVis-IM18 and GraSP-IM dataset.}
    \label{fig:vis}
\end{figure}
\subsection{State-of-the-art Comparison}
In this work, we compare to prior state-of-the-art methods from both the surgical domain~\cite{wang2024video} and the general computer vision domain~\cite{yan2024visa,rong2025mpg}. VIS-Net~\cite{wang2024video} is a surgery-specific model that combines spatial-temporal graph features with language-guided attention. VISA~\cite{yan2024visa} and MPG-SAM 2~\cite{rong2025mpg} are general-purpose frameworks that use large language models and SAM-based propagation for referring video object segmentation. As shown in Table~\ref{tab1}, even though prior works emphasize vision-language alignment and temporal modeling, our proposed approach consistently outperforms them across multiple datasets, e.g., achieving 89.91 J on EndoVis-IM17 compared to 88.2 by MPG-SAM 2, and 84.48 J\&F on EndoVis-IM18 compared to 83.03. This demonstrates our model’s superior ability to fuse motion-guided expressions with language and video cues, enabling more precise spatial-temporal localization. As shown in Figure~\ref{fig:vis}, our model produces accurate spatial-temporal segmentations in challenging scenarios.
\subsection{Generalizibility Analysis}
While prior methods demonstrate the benefits of vision-language alignment and temporal modeling, Table~\ref{tab1} shows that our approach not only achieves state-of-the-art performance on standard surgical benchmarks (EndoVis-IM17/18) but also exhibits strong zero-shot generalization across different surgical procedures and modalities. Specifically, our model, trained under a unified setting without any dataset-specific tuning, generalizes effectively to CholecSeg8k-IM (laparoscopic cholecystectomy) using the model trained on GraSP-IM (robot-assisted prostatectomy), both of which differ significantly from the training data in terms of anatomy, tool types, imaging modalities, and institutional sources. The consistent performance gains on the unseen dataset highlight the robustness of our motion-guided language grounding framework, which captures instrument behavior beyond static appearance or naming cues, making it broadly applicable to diverse clinical environments.
% our model produces accurate spatial-temporal segmentations even in challenging scenarios involving occlusion, tool clutter, and visually ambiguous contexts, confirming its ability to generalize motion-centric expressions across domains.
\begin{table}[htbp]
\centering
\resizebox{\linewidth}{!}{
\begin{tabular}{llcccccc}
\toprule
\multirow{2}{*}{Dataset} & \multirow{2}{*}{Expr.} 
& \multicolumn{3}{c}{w/o Motion} 
& \multicolumn{3}{c}{w/ Motion} \\
\cmidrule(lr){3-5} \cmidrule(lr){6-8}
& & J & F& J\&F & J & F & J\&F \\
\midrule

\multirow{3}{*}{EV-IM17} 
& Appr.          & 73.92 & 74.58  & 74.25 & 85.42 & 84.91 & 85.17 \\
& Spatial  & 76.33 & 76.95    & 76.64 & 87.52 & 87.13 & 87.33 \\
& Motion & 79.92 & 78.14    & 79.03   & 89.91 & 88.93 & 89.42 \\
\midrule

\multirow{3}{*}{EV-IM18} 
& Appr. & 67.78 & 68.53 & 68.16 & 82.77 & 79.26 & 81.02 \\
& Spatial       & 69.24 & 69.05 & 69.15 & 84.13 & 81.97 & 83.05 \\
& Motion & 71.79 & 70.32 & 71.06 & 85.91 & 83.05 & 84.48 \\
\midrule

\multirow{3}{*}{GSP-IM} 
& Appr. & 68.01 & 68.75 & 68.38 & 81.02 & 81.93 & 81.48 \\
& Spatial      & 68.92 & 69.88 & 69.40 & 81.53 & 82.74 & 82.14 \\
& Motion & 70.18 & 71.34 & 70.76 & 84.92 & 85.01 & 84.97 \\
\bottomrule

\end{tabular}}
\caption{Evaluation on different testing expressions (Expr.): appearance-based instrument descriptions (Appr.), instrument spatial information-based expressions (Spatial), and motion-aware temporal dynamics expressions (Motion). Performance is compared between models trained with (w/) and without (w/o) motion-aware expressions.}
\label{tab3}
\end{table}
\subsection{Motion-guided Expression Analysis}
To evaluate the impact of motion-centric supervision, we compare models trained with and without motion-based expressions across different test-time referring styles: (1) appearance-based instrument descriptions  (i.e., describing the name, appearance, and function of a given instrument), (2) instrument spatial information-based expressions (e.g., ``on the left''), and (3) full motion-based descriptions (e.g., ``the tool entering from the right and grasping the gallbladder''). As shown in Table~\ref{tab3}, models trained with motion-aware expressions consistently outperform those trained without, across all datasets and expression styles. For example, on EndoVis-IM17, SurgRef trained with motion expressions achieves a J\&F of 89.42, compared to 79.03 for non-motion-trained counterparts. Even when tested on expression with static features including appearance-based and spatial-based expressions, motion-trained models show substantial gains (e.g., 85.17 vs. 74.25 on appearance-based expressions). This consistent improvement confirms that motion-based supervision from our Ref-IMotion dataset enhances not only motion expression understanding but also generalizes to static feature descriptions, demonstrating stronger robustness in real-world surgical contexts.
% \begin{table}[htbp]
% \centering
% \small
% \resizebox{\linewidth}{!}{%
% \begin{tabular}{llccc}
% \toprule
% \textbf{Evaluation} & \textbf{Training Expr. $\rightarrow$ Testing Expr.} & \textbf{J} & \textbf{F} & \textbf{J\&F} \\
% \midrule
% Same-style 
% & origin $\rightarrow$ origin            & 87.11 & 88.29 & 87.70 \\
% & no location $\rightarrow$ no location & 59.24 & 69.02 & 64.13 \\
% & no name $\rightarrow$ no name         & 78.05 & 79.98 & 79.02 \\
% \midrule
% Cross-style 
% & origin $\rightarrow$ no location      & 57.78 & 61.54 & 59.66 \\
% & origin $\rightarrow$ no name          & 62.24 & 64.23 & 63.23 \\
% & no location $\rightarrow$ origin      & 59.24 & 69.07 & 64.15 \\
% & no location $\rightarrow$ no name     & 34.79 & 42.50 & 38.64 \\
% & no name $\rightarrow$ origin          & 78.62 & 80.69 & 79.65 \\
% & no name $\rightarrow$ no location     & 41.42 & 43.84 & 42.63 \\
% \bottomrule
% \end{tabular}
% }
% \caption{Ablation results (\%) on EndoVis-IM17 under different expression styles (Expr.: Expression). Same-style and cross-style evaluations for training and testing are shown.}
% \label{tab2}
% \end{table}
\begin{table}[htbp]
\centering
\resizebox{\linewidth}{!}{
\begin{tabular}{llccc}
\multicolumn{5}{c}{\textit{(a) Expression style ablation}} \\
\toprule
\text{Evaluation} & Training $\rightarrow$ Testing & J & F & J\&F \\
\midrule
Same-style 
& origin $\rightarrow$ origin            & 87.11 & 88.29 & 87.70 \\
& no location $\rightarrow$ no location & 59.24 & 69.02 & 64.13 \\
& no name $\rightarrow$ no name         & 78.05 & 79.98 & 79.02 \\
\midrule
Cross-style 
& origin $\rightarrow$ no location      & 57.78 & 61.54 & 59.66 \\
& origin $\rightarrow$ no name          & 62.24 & 64.23 & 63.23 \\
& no location $\rightarrow$ origin      & 59.24 & 69.07 & 64.15 \\
& no location $\rightarrow$ no name     & 34.79 & 42.50 & 38.64 \\
& no name $\rightarrow$ origin          & 78.62 & 80.69 & 79.65 \\
& no name $\rightarrow$ no location     & 41.42 & 43.84 & 42.63 \\
\bottomrule
\end{tabular}}
\resizebox{\linewidth}{!}{
\begin{tabular}{lccccc}
\multicolumn{6}{c}{\textit{(b) Key-frame selection strategy and T' ablation}} \\
\toprule
Selection Strategy & T'= 4 & T'= 8 & T'= 16 & T'= 24 & 
T'= 32\\
\midrule
Uniform sampling & 79.28 & 82.16 & 84.67& 85.92& 86.77\\
Cosine similarity & 78.93 & 85.71 & 86.82 & 86.95& 87.09\\
Ours & 80.33 & 89.42 & 89.56 & 88.87& 88.35\\
\bottomrule
\end{tabular}}
\caption{Ablation study on EndoVis-IM17 (\%). (a) Expression style generalization with the same style and cross-style for training expression and testing expression. (b) Key-frame selection strategies across different T' values (J\&F scores).}
\label{tab2}
\end{table}

\subsection{Ablation Study}
\paragraph{Expression Style.}
Table~\ref{tab2} presents a detailed analysis of model's performance under varying referring expression styles on EndoVis-IM17. In the same-style setting, where the model is both trained and tested on the same expression type, performance is highest for original expressions (J\&F = 87.70), while removing location or name cues leads to substantial drops, especially without spatial cues (J\&F = 64.13), confirming the utility of such information. In the cross-style setting, performance degrades further when there's a mismatch between training and testing styles. For example, the origin-trained model tested on expressions with no location yields a J\&F of 59.66, compared to 87.70 when tested on origin expressions. Notably, the model trained without instrument names still generalizes well to origin expressions (J\&F = 79.65), highlighting that motion and spatial context can compensate for missing lexical identifiers. Conversely, models trained without location cues generalize poorly to other styles, particularly when both name and location cues are absent (J\&F = 38.64 or 42.63). Overall, these results emphasize that while spatial and naming cues are helpful, motion-centric training enhances generalization, especially under challenging linguistic conditions.
\paragraph{Key-Frame Selection.}
As shown in Table~\ref{tab2}, our object-level key-frame selection achieves optimal J\&F of 89.42 at $T'=8$, demonstrating superior sample efficiency. The performance saturation while $T'$ increases indicates that our method effectively identifies the most expression-relevant frames, while additional frames introduce significant training cost without improving performance.
In contrast, uniform sampling and Cosine similarity-based selections exhibit monotonic improvement, requiring more frames to capture task-relevant temporal content.

\section{Conclusion}
In this work, we introduced SurgRef, a motion-guided referring video segmentation framework that grounds natural language expressions in surgical videos by explicitly modeling instrument motion. Paired with our curated Ref-IMotion dataset, comprising motion-centric expressions and dense spatial-temporal annotations, SurgRef advances beyond static appearance-based methods by capturing dynamic tool behaviors and semantic trajectories. Our model achieves state-of-the-art performance across multiple surgical benchmarks and demonstrates strong zero-shot generalization to unseen procedures, tools, and modalities. Together, SurgRef and Ref-IMotion form a strong foundation for motion-aware, language-driven surgical understanding.

\section{Acknowledgments}
Meng Wei is supported by the UKRI EPSRC CDT in Smart Medical Imaging [EP/S022104/1]. Tom Vercauteren is a co-founder and shareholder of Hypervision Surgical. This work is supported by core funding from the Wellcome Trust / EPSRC [WT203148/Z/16/Z; NS/A000049/1] and has received funding from the European Union (ERC, CompSURG, 101088553). Views and opinions expressed are however those of the authors only and do not necessarily reflect those of the European Union or the European Research Council. Neither the European Union nor the granting authority can be held responsible for them. This work was also partially supported by French state funds managed within the Plan Investissements d'Avenir by the ANR under reference ANR-10-IAHU-02 (IHU Strasbourg).

\bibliography{aaai2026}

\end{document}